\documentclass[10pt,twocolumn,letterpaper]{article}
\usepackage[pagenumbers]{wacv}

\usepackage[utf8]{inputenc} 
\usepackage[T1]{fontenc}    
\usepackage{hyperref}       
\usepackage{url}            
\usepackage{booktabs}       
\usepackage{amsfonts}       
\usepackage{nicefrac}       
\usepackage{microtype}      
\usepackage{graphicx}
\usepackage[numbers, sort]{natbib}
\usepackage{doi}

\usepackage{amsmath}
\usepackage{amssymb}
\usepackage{mathtools}
\usepackage{amsthm}

\usepackage{multirow}
\usepackage{stfloats}

\RequirePackage{algorithm}
\RequirePackage{algorithmic}

\usepackage{microtype}
\usepackage{graphicx}
\usepackage{booktabs} 

\usepackage{wrapfig}
\usepackage{caption}
\usepackage[capitalise]{cleveref}

\theoremstyle{plain}

\theoremstyle{definition}

\theoremstyle{remark}

\usepackage{setspace}

\title{Is Self-Supervised Pre-training on Satellite Imagery Better than ImageNet? A Systematic Study with Sentinel-2}

\author{%
  Saad Lahrichi\\
  University of Missouri\\
  Columbia, MO 65201\\
  \texttt{\small saad.lahrichi@missouri.edu} \\
  \and
  Zion Sheng \\
  Duke University\\
  Durham, NC 27708 \\
  \texttt{ \small zion.sheng@duke.edu} \\
  \and
  Shufan Xia \\
  Stanford University\\
  Stanford, CA 94305 \\
  \texttt{\small sxia11@stanford.edu} \\
  \and
  Kyle Bradbury \\
  Duke Universiry \\
  Durham, NC 27708\\
  \texttt{\small kyle.bradbury@duke.edu} \\
  \and
Jordan Malof\\
University of Missouri\\
Columbia, MO 65201\\
{\tt \small jmdrp@missouri.edu}
}

\date{}

\hypersetup{
pdftitle={Is Self-Supervised Pre-training on Satellite Imagery Better than ImageNet? A Systematic Study with Sentinel-2},
pdfsubject={cs.LG, cs.AI},
pdfauthor={Saad Lahrichi, Zion Sheng, Shufan Xia, Kyle Bradbury, Jordan Malof },
pdfkeywords={self-supervised learning, satellite imagery, ImageNet, pretraining, Sentinel},
}

\begin{document}
\maketitle

\begin{abstract}
  Self-supervised learning (SSL) has demonstrated significant potential in pre-training robust models with limited labeled data, making it particularly valuable for remote sensing (RS) tasks. A common assumption is that pre-training on domain-aligned data provides maximal benefits on downstream tasks, particularly when compared to ImageNet-pre-training (INP). In this work, we investigate this assumption by collecting GeoNet, a large and diverse dataset of global optical Sentinel-2 imagery, and pre-training SwAV and MAE on both GeoNet and ImageNet. Evaluating these models on six downstream tasks in the few-shot setting reveals that SSL pre-training on RS data offers modest performance improvements over INP, and that it remains competitive in multiple scenarios. This indicates that the presumed benefits of SSL pre-training on RS data may be overstated, and the additional costs of data curation and pre-training could be unjustified. 

\end{abstract}


\section{Introduction}
\label{sec:intro}
Data-driven models, such as deep neural networks (DNNs), have been found to be very successful for recognition of remote sensing (RS) data, such as optical or multispectral satellite imagery \cite{bai2023deep, joshi2023remote, adegun2023review}.  However, one major limitation of DNN-based approaches is their need for large quantities of annotated training data.  One widespread approach to reduce the reliance on labeled training data is to pre-train the model on a related task with abundant training data availability.  A common choice has been to pre-train on the ImageNet dataset using supervised learning.  In this setting, the model or its weights are said to be \textit{pretrained} on a \textit{source task} (e.g., ImageNet), and then \textit{fine-tuned} on a \textit{downstream task} (e.g., a RS recognition task).   

Recent research has suggested that pre-training with self-supervised learning (SSL) on ImageNet produces better downstream performance compared to supervised learning on ImageNet \cite{ericsson2021well}. This finding has been corroborated for downstream tasks that involve RS data \cite{calhoun2022self}.  In contrast to supervised learning, SSL does not require ground-truth annotations, enabling SSL pretraining on very large datasets that lack abundant annotations, such as RS data.  Pre-training is generally assumed to be most effective when the source and downstream data are most similar \cite{hammoud2024pretraining, he2022masked}, a concept we term here as \textit{ domain-aligned pre-training}. This assumption has motivated substantial recent research focused on SSL pretraining on RS datasets (RSP), instead of ImageNet SSL pre-training (INP) \cite{calhoun2022self, manas2021seasonal, ayush2021geography, xiong2024earthnets, cong2022satmae}.  This research has included both the development of specialized SSL methods for RSP \cite{jean2019tile2vec,manas2021seasonal,ayush2021geography,cong2022satmae}, as well as the creation of large RS datasets designed to support RSP \cite{wang2023ssl4eo, tao2023tov, bastani2023satlaspretrain}. 

These studies usually report improvements in downstream accuracy compared to previous methods. However, a common limitation of recent work is the inconsistent use of INP as a baseline for comparison.  This is a reasonable choice since it is generally assumed that domain-aligned pretraining is superior, which is supported by recent studies showing that RSP outperforms INP on downstream RS tasks.  Yet, other studies have reported that INP remains competitive with RSP \cite{corley2024revisiting, wang2022empirical}, challenging these assumptions. If INP proves to be competitive with RSP, it carries significant theoretical and practical implications, highlighting the importance of this question. Practically, the vision community regularly pre-trains large models on ImageNet, while building analogous pre-trained models on RS data requires substantial additional compute and effort: is this cost and the associated research effort justified? 

\textbf{In this work, we investigate whether SSL pre-training on RS data offers a genuine advantage over ImageNet pre-training, when used for downstream RS tasks.}  To answer this question, we systematically compare RSP to INP on various downstream RS tasks. To maintain a tractable scope for our study, we focus on Sentinel-2, comprising medium-resolution optical satellite imagery. Sentinel-2 has received substantial attention in recent years, particularly for SSL pre-training \cite{wang2023ssl4eo, manas2021seasonal, leenstra2021self, yuan2022sits, scheibenreif2022self}, due to its wide geographic and temporal availability.  Leveraging this feature, we constructed GeoNet, a large, diverse, global collection of optical Sentinel-2 imagery.  We pre-trained large vision models on GeoNet and ImageNet, respectively, and evaluated their performance when fine-tuned on six different downstream recognition tasks involving Sentinel-2 imagery (except one).  To ensure fairness, we conducted the pre-training ourselves using identitical procedures, rather than downloading pre-trained checkpoints (e.g., for the ImageNet model). Building on recent work \cite{isprs-archives-XLIII-B3-2022-1399-2022}, we also consider a two-stage approach, where we pre-train first on ImageNet, and then GeoNet.

For each pre-training dataset, we pre-train using two SSL techniques: swapping assignments from multiple views (SwAV)  \cite{caron2020unsupervised}, and masked auto-encoders (MAE) \cite{he2022masked}. SwAV is representative of contrastive SSL approachces, while MAE is representative of generative approaches. Our primary goal is to evaluate the efficacy of domain-aligned \textit{data}; therefore, we utilize generic SSL methods rather than specialized RSP models (e.g., SeCo \cite{manas2021seasonal}, SatMAE \cite{cong2022satmae}, Tile2Vec \cite{jean2019tile2vec}, etc).  Surprisingly, our experiments suggest that there is only a modest advantage of RSP, prompting us to conduct several analyses to explain these findings. Notably, we compare the efficacy of GeoNet with another recently proposed large dataset of Sentinel-2 imagery specifically tailored for SSL: SSL4EO \cite{wang2023ssl4eo}.  We also investigate whether RSP leads to better feature representations for the data in our downstream tasks than INP, and whether the efficacy of the feature representations correlates well with recognition accuracy.  We summarize the contributions of this work as follows: 
\begin{itemize}

\item We develop GeoNet, a large and diverse dataset of optical Sentinel-2 imagery, curated specifically for SSL in remote sensing.

\item Using GeoNet, we systematically compare pre-training on remote sensing data with pre-training on ImageNet, using two state-of-the-art self-supervised learning (SSL) techniques: SwAV and MAE.

\item We find that pre-training with remote sensing data does not offer consistent improvements in downstream performance as compared to pre-training with ImageNet, and we conduct analyses to explain this finding.  

\end{itemize}

The remainder of this paper is organized as follows: \cref{sec:related-works} reviews related work, \cref{sec:methods} describe datasets and methods adopted for our experiments, \cref{sec:geonet_description} introduces GeoNet, \cref{sec:experiments} details our experiments, \cref{sec:results} presents our results, \cref{sec:discussion} analyzes the results, and \cref{sec:conclusions} summarizes our conclusions.

\section{Related Works}
\label{sec:related-works}

\paragraph{Self-Supervised Learning for Pre-Training}
A large number of studies have reported that SSL pre-training is competitive with, or outperforms, supervised learning models\cite{caron2020unsupervised, misra2020self,chen2020simple,he2020momentum, he2022masked}.  Early SSL approaches involved solving various pretext tasks such as colorization \cite{zhang2016colorful}, relative patch prediction \cite{doersch2015unsupervised}, rotation prediction \cite{gidaris2018unsupervised}, jigsaw puzzle solving \cite{noroozi2016unsupervised}, and image inpainting \cite{pathak2016context}. State-of-the-art SSL generally adopt one of two general strategies: contrastive or generative.  In contrastive approaches, models are trained to discriminate between positive and negative pairs and learn distinct representations.  Example methods include SimCLR \cite{chen2020simple}, MoCo \cite{he2020momentum, chen2020mocov2}, BYOL \cite{grill2020bootstrap}, and SwAV \cite{caron2020unsupervised}.  We adopt SwAV for our experiments due to its state-of-the-art performance, extensive use in previous research \cite{isprs-archives-XLIII-B3-2022-1399-2022, hakizimana2024enhanced, calhoun2022self, guldenring2021self}, and good transfer performance \cite{ericsson2021well}. Generative approaches, on the other hand, learn meaningful representations by masking patches in the input images and reconstructing the missing pixels. Popular generative methods include SimMIM \cite{xie2022simmim}, BEiT \cite{bao2021beit}, iGPT \cite{chen2020generative}, and Masked Autoencoders (MAE) \cite{he2022masked}. We adopt MAE for our experiments due to its state-of-the-art performance, scalability to large datasets, fast training speed, and widespread adoption \cite{cong2022satmae, tang2024cross, reed2023scale, lin2023ss}.

\paragraph{SSL for Remote Sensing}
RS data exhibit unique properties compared to natural imagery, which has motivated the development of RSP-specific methods, as well as large RS datasets to support domain-aligned RSP. A large number of publications have focused on developing RSP SSL methods, e.g. Tile2Vec \cite{jean2019tile2vec}, Seasonal Contrast (SeCo) \cite{manas2021seasonal}, Geography-Aware Self-Supervised Learning \cite{ayush2021geography}, and SatMAE \cite{cong2022satmae}.  We refer readers to \cite{wang2022self, tao2023self} for a more detailed discussion of SSL methods for RS. While these models leverage the temporal and multi-spectral features of RS data, this work focuses on evaluating the benefits of domain-aligned \textit{data}: i.e., whether using RS data offers advantages over natural imagery, when applying the same SSL method. Therefore, we restrict our analysis to models suitable for RGB images which can be pretrained on both GeoNet and ImageNet.  Multiple datasets have been designed for both supervised \cite{cornebise2022open,christie2018functional,lam2018xview,demir2018deepglobe,gupta2019creating,helber2019eurosat,johnson2022opensentinelmap,tao2023tov,bastani2023satlaspretrain,xiong2024earthnets} and SSL pre-training on satellite imagery \cite{shen2023firerisk, li2021geographical}.  Our study focuses on Sentinel-2 optical imagery, as it is the highest resolution imagery available globally, with high revisit rates, and for free. To use Sentinel-2 imagery for SSL, researchers have developed different methods for dataset curation. The SeCo dataset \cite{manas2021seasonal} comprises 1 million images from 200K locations, with sampling focused around cities. The dataset includes 5 images per location, each taken 3 months apart. The SSL4EO dataset \cite{wang2023ssl4eo} improved upon SeCo by removing overlapping imagery, and emphasizing geographic diversity, with 250K locations with 4 images per location, also totaling 1 million optical images.  In this work, we constructed GeoNet, which comprises 1 million images, emphasizing geographic diversity more than prior datasets, with the intent of maximizing the geographic robustness of models trained on it. The details of GeoNet are presented in \cref{sec:geonet_description}.

\paragraph{Effectiveness of Domain-Aligned Pre-Training}
The benefits of domain-aligned pre-training for RS vision tasks remains inconclusive due to mixed conclusions about it in prior studies, and inconsistent or improper comparisons with INP as a baseline. For scene classification tasks, \cite{dimitrovski2024domain} found that RSP consistently, albeit narrowly, outperforms INP across tasks and transfer learning methods, including linear probing and fine-tuning. \cite{manas2021seasonal, wang2023ssl4eo, ayush2021geography} reached similar conclusions, although they generally used supervised  training on ImageNet as a baseline rather than INP. In contrast, other research found that the benefits of RSP are less evident. The results in \cite{isprs-archives-XLIII-B3-2022-1399-2022} suggest that RS pre-training can be beneficial for scene classification, especially when the pre-training and downstream datasets share common classes. However, INP remained competitive, performing best in the fine-tuning scenario. They achieved state-of-the-art results by using domain-adaptive pre-training, a two-stage SSL process, pre-training first on ImageNet and then on RS data. \cite{zhang2022consecutive} expanded on this idea and proposed Consecutive PreTraining, which involves performing INP then RSP. They found that it outperformed both supervised ImageNet and supervised in-domain pre-training, but did not compare to either INP or RSP. \cite{corley2024revisiting} found that for classification tasks, INP remains a more competitive baseline than previously thought, particularly when using consistent preprocessing. \cite{wang2022empirical} found that while RSP can be beneficial, the performance improvements depend significantly on the task and level of detail required. Notably, most of these studies use high-resolution RS datasets (e.g. Million-AID) for RSP, and many recommend that future work should explore the development of a large RS dataset for SSL. In this work, we build upon these studies by evaluating the benefits of in-domain data for SSL pre-training (specifically using Sentinel-2 imagery), as opposed to out-of-domain SSL pre-training (i.e. ImageNet), and we do so across both classification and the less-studied segmentation tasks.

\section{Materials and Methods}
\label{sec:methods}

\subsection{Benchmark Datasets}
To evaluate the advantages of RSP over INP and the impact of a domain-adaptive strategy, we compile a list of accessible, well-studied, and highly representative benchmark datasets for two categories of downstream tasks: classification and semantic segmentation. We guide our selection by prioritizing datasets that include Sentinel-2 imagery, aligning with the data used for curating GeoNet. We also consider benchmark datasets with resolutions ranging from 0.1 to 30 m/px to evaluate model performance on tasks at different resolutions from the original training data. We further assess the geographic diversity of the datasets to evaluate GeoNet's generalizability across regions. Based on these criteria, our final selection of benchmark datasets includes SEN12MS, DeepGlobe, Farm Parcel Delineation, BigEarthNet, EuroSAT, and LandCoverNet. \cref{tab:benchmarks} summarizes the key features of these datasets, and the Appendix has detailed descriptions of each dataset and our preprocessing.

\begin{table}[]
\centering
\resizebox{\columnwidth}{!}{%
\begin{tabular}{lcccc}
\toprule
\textbf{Dataset} & \textbf{Task} & \textbf{Resolution (size)} & \textbf{Classes} & \textbf{Region} \\ 
\midrule
SEN12MS \cite{schmitt2019sen12ms}  & Segmentation   & 10 m/px (224×224)     & 10 & Global \\ 
DeepGlobe \cite{demir2018deepglobe} & Segmentation  & 0.5 m/px (224×224)    & 7  & Asia \\ 
Field Delineation \cite{aung2020farm} & Segmentation & 10 m/px (224×224) & 1 & France \\ 
BigEarthNet \cite{sumbul2019bigearthnet} & Multi-label Cls. & 10 m/px (120×120)  & 43  & Europe\\ 
EuroSAT \cite{helber2019eurosat}& Multi-class Cls. & 10 m/px (224×224) & 10 & Europe \\
LandCoverNet \cite{alemohammad2020landcovernet} & Segmentation & 10 m/px (256×256) & 4 & Africa \\ 
\bottomrule
\end{tabular}
}
\caption{Benchmark datasets used in our experiments}
\label{tab:benchmarks}
\end{table}

\subsection{Self-Supervised Learning Methods}
\label{sub:ssl_models}
\textbf{SwAV} \cite{caron2020unsupervised} combined ideas from contrastive learning and clustering, and trained models to assign feature representations from one view of an image to the cluster obtained from another view. By swapping these cluster assignments, this contrastive learning approach eliminates the need for large batch sizes or memory banks, while outperforming competitor models. 

\textbf{MAE} \cite{he2022masked} is a generative SSL method that learns representations by masking random patches in the input image and training the model to reconstruct them. The reconstruction error is used to update the model's weights. Two main hyperparameters can be set: patch size and masking ratio. The former (default set to 16) controls the size of masked patches while the latter (default set to 0.75) determines the amount of patches masked that the model learns to fill. Larger patches and higher ratios make the reconstruction task more challenging.

\section{The GeoNet Dataset}
\label{sec:geonet_description}

GeoNet comprises approximately 1 million Sentinel-2 optical images, collected using Google Earth Engine \footnote{Code used to download the data can be found in \hyperlink{https://github.com/zcalhoun/data-plus-22}{this repository}}. GeoNet uses a novel sampling strategy that places greater emphasis on geographic diversity and RS task relevance than other existing datasets for Sentinel-2 (see \cref{sec:related-works}), with the goal of producing pre-trained models that transfer better across different RS tasks and regions.

\begin{figure}
    \centering
    \includegraphics[width=\linewidth]{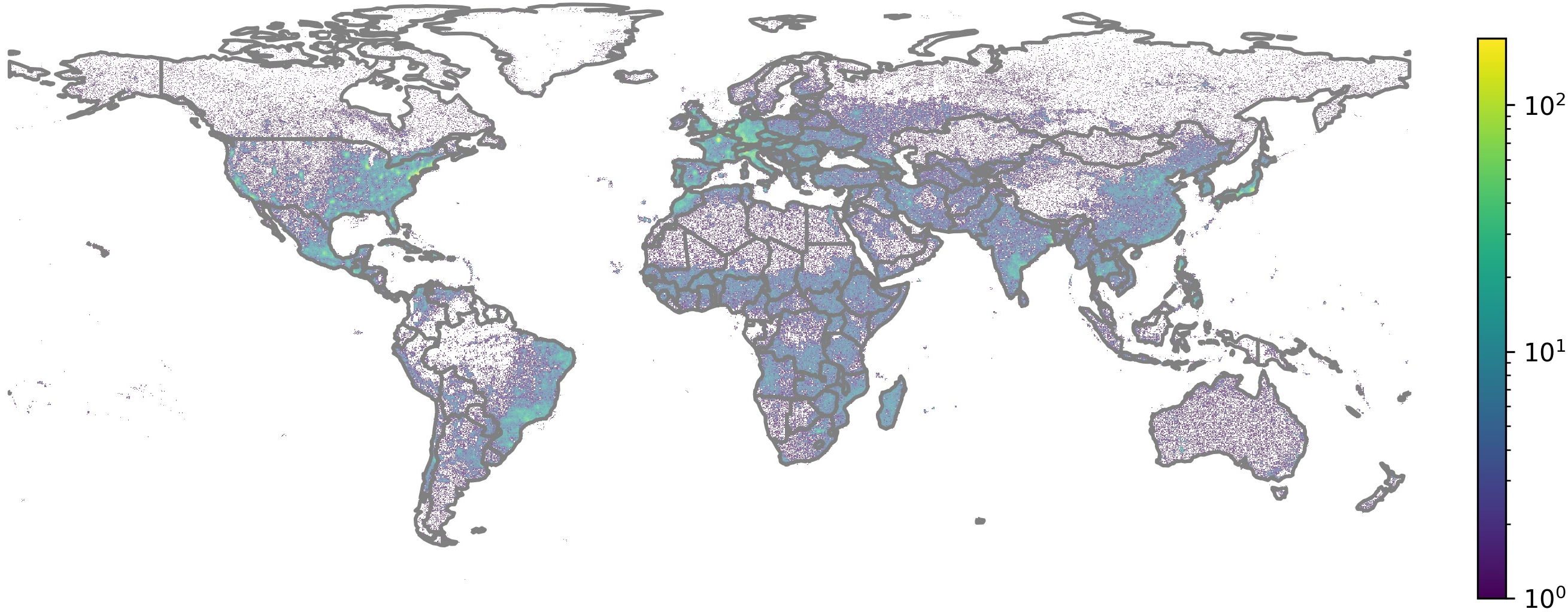}
    \caption{GeoNet images distribution. Colors represent the number of images per 22 km$^2$ area
}
    \label{fig:geonet_map}
\end{figure}

\begin{figure}
    \centering
    \includegraphics[width=\linewidth]{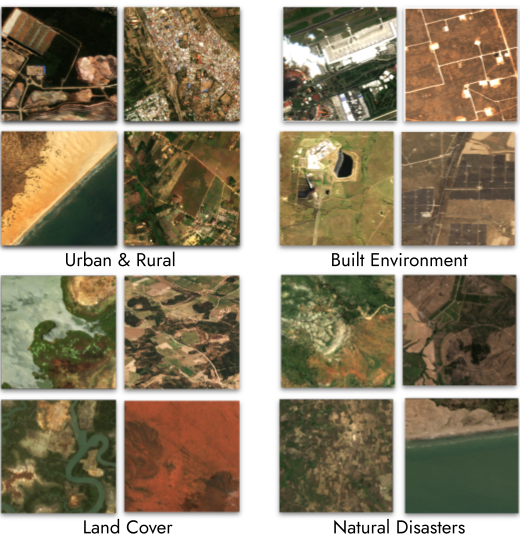}
    \caption{Example imagery from GeoNet using each of the sampling strategies}
    \label{fig:geonet_examples}
\end{figure}

We summarize the design of GeoNet here, but full details can be found in \cref{supp:sampling}. To build GeoNet, we divide the earth's surface into a grid, where each grid cell corresponds to 224x224 pixels in size: the size of an input image to our SSL models.  Each grid cell is then treated as a candidate image that can be included in GeoNet.  Our objective is to build GeoNet to be similar in size and modalities (i.e., image bands) to ImageNet, while maximizing the geographic diversity and relevance to common vision tasks in RS. To accomplish this goal, we sample grid cells at random across the globe, but while adjusting the sampling probability of cell based upon its likely relevance to common RS tasks.  We then utilize publicly-available data as proxies for the degree of RS task relevance for each grid cell. We organize these proxy data into four broad categories, and we randomly sample a fixed proportion of the GeoNet dataset based upon each criterion.  We next describe each of these four criterion, and the total proportion of GeoNet data that was sampled using it.  

\textbf{Urban and Rural Regions (60\%):} This category is subdivided into two: urban (40\%) and rural (20\%).  For urban areas, we adapt the approach in \cite{manas2021seasonal, wang2023ssl4eo} by first sampling a city, with the probability of sampling any city proportional to $log(\text{population})$. We then draw a random sample coordinate from a 2-dimensional (i.e. spatial) normal distribution centered on that city, with a standard deviation of 50 km radius. For "rural" areas, we identify all global grid cells with population density between 5 and 250 people/km${^2}$ and then randomly sample among these locations, without replacement.

\textbf{Land Cover Types ($\approx$20\%):} A total of 2\% of cells is allocated to each of eleven land cover types: artificial land, cropland, grassland, tree-covered areas, shrub-covered areas, herbaceous vegetation (aquatic or regularly flooded), mangroves, sparse vegetation, bare soil, snow and glaciers, and water bodies. No samples are collected for water bodies, as limited information can be extracted from RGB images of these areas. 

\textbf{Natural Disasters ($\approx$19\%):} We identify locations that are vulnerable to, or have a history of, natural disasters: e.g. wildfires, cyclones, droughts, and floods. We then sample these locations randomly, with greater probabilities assigned to locations with greater disaster severity or occurrence frequency. 

\textbf{Built Environment (<1\%):} We used public data sources to identify geo-locations of seven human-built features: airports, mining sites, power plants, ports, oil rigs, wind turbines, and dams. Our search returned approximately 82k coordinates, and all of these locations were included in GeoNet.

\cref{fig:geonet_examples} presents imagery sampled according to each of the proxy categories discussed above, illustrating its diversity.  \cref{fig:geonet_map} presents the distribution of sampled images that results from this sampling process, which reveals global coverage, but with significant differences in the degree of representation.  For example, there is relatively limited sampling in Siberia or Northern Canada, due to the limited task relevance and high visual redundancy of the scenes there.

\section{Experiments}
\label{sec:experiments}

\begin{figure}
    \centering
    \includegraphics[width=\linewidth]{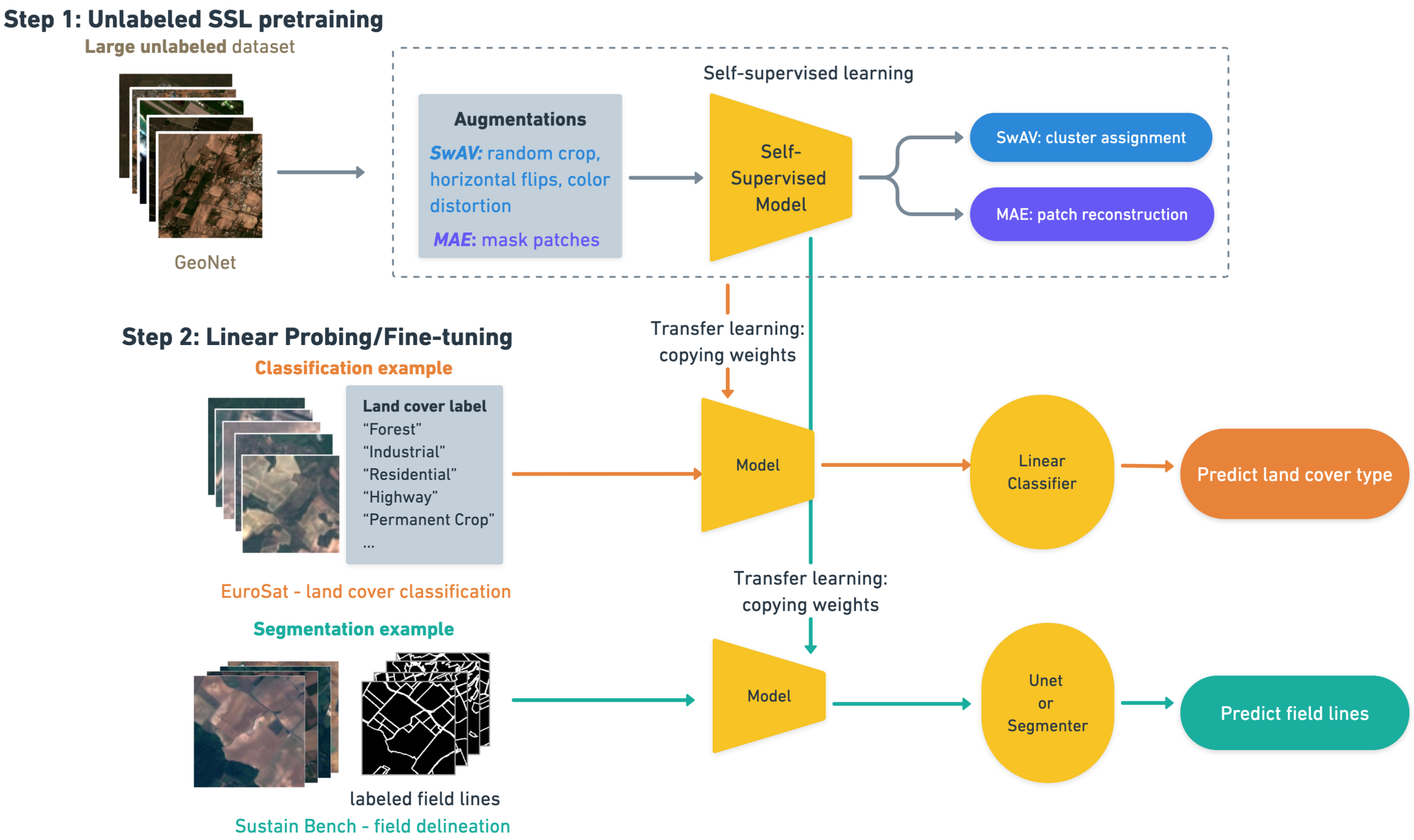}
    \caption{Diagram representing our experimental setup. We first train SwAV and MAE on GeoNet, then evaluate their downstream performance on several tasks.}
    \label{fig:exp_setup}
\end{figure}

The primary goal of our experiments is to evaluate whether RSP generally yields better downstream performance than INP, when controlling for as many experimental factors as possible, except for the source of the pre-training data (i.e., Sentinel-2 or ImageNet).  To address this question, we pre-train models on ImageNet and GeoNet and compare their performance on six downstream RS tasks. We repeat this comparison for two pre-training strategies: SwAV and MAE. Our overall experimental design is illustrated in \cref{fig:exp_setup}.  

\paragraph{Pretraining} The precise details of the pre-training design  (e.g., epochs trained, data preprocessing, etc.) are adopted from the SSL literature as much as possible, especially \cite{caron2020unsupervised} and \cite{he2022masked} that introduced SwAV and MAE, respectively.  The precise experimental details are provided in the supplement, but we highlight a few key features here. Models pre-trained on GeoNet and ImageNet are preprocessed in the same manner, and then allocated precisely the same quantity of computation.  Specifically, the size of the images ingested by the SSL algorithms are identical, and they are trained for the same number of epochs.  Notably, these images are cropped from IN and GN, respectively, and then resized to be the same size (i.e., 224x224).  Because ImageNet images are, on average, larger than GeoNet images, the quantity of unique pixels used to pre-train ImageNet is larger (by about four times, on average). Therefore, although the compute is controlled, ImageNet may have an advantage due to the availability of more source pixels.  We hypothesize that this advantage is countervailed somewhat by two factors: (i) the high similarity of pixels taken from within the same images, reducing the effective uniqueness of the data (e.g., as compared to utilizing more unique images); (ii) and the fact that pixels near the edges (which contain most of the extra pixels) are less likely to be sampled by the cropping procedure. Nonetheless, this is an important limitation of our experimental design.

\paragraph{Downstream Tasks} We evaluate the efficacy of each pre-trained model using linear probing (LP) and end-to-end finetuning (FT) on our downstream tasks (see \cref{sub:benchmark_datasets} for tasks), as illustrated in \cref{fig:exp_setup}.  The efficacy of different pre-training strategies tends to be greatest when there is limited downstream training data; this is also the same scenario when pre-training is most practically useful. Therefore, we conduct experiments with relatively limited quantities of labeled data: 64, 256, or 1024 labeled instances, respectively.  For each downstream task, we append an adequate decoder: Unet/Segmenter for segmentation tasks, or a fully connected Linear layer for classification tasks.  We train each downstream model for 100 epochs, using the Adam optimizer. The learning rate was initially set at $1e-3$, then manually adjusted to ensure convergence of each encoder/task combination if the initial setting did not result in consistent convergence. LP and FT for each pre-trained model, benchmark task, and training data size is repeated five times, following the strategy of \cite{calhoun2022self}. At test time, we use the model with the lowest validation loss \footnote{Except for BigEarthNet, where we use the highest validation accuracy, as we found it to be uncorrelated to the loss for this dataset only} and report the average accuracy metric across the five runs. For semantic segmentation, we use the Jaccard loss and report the micro IoU, while cross-entropy loss and micro F-1 score are used for multi-class and multi-label classification. We use 256 samples for validation and 2048 samples for testing (or the entire val/test set if smaller). \cref{fig:exp_setup} describes the experimental setup.

\section{Results}
\label{sec:results}

\cref{tab:few_shot_lin_prob_ft} summarizes the main results of our experiments. The top half of the table shows linear probing results, and the bottom half presents the fine-tuning ones. The table report the accuracies of MAE-GN, MAE-IN, SwAV-GN, SwAV-IN, and Sup.-IN across three training sizes and six benchmark tasks.

\begin{figure}
    \centering
    \includegraphics[width=0.9\linewidth]{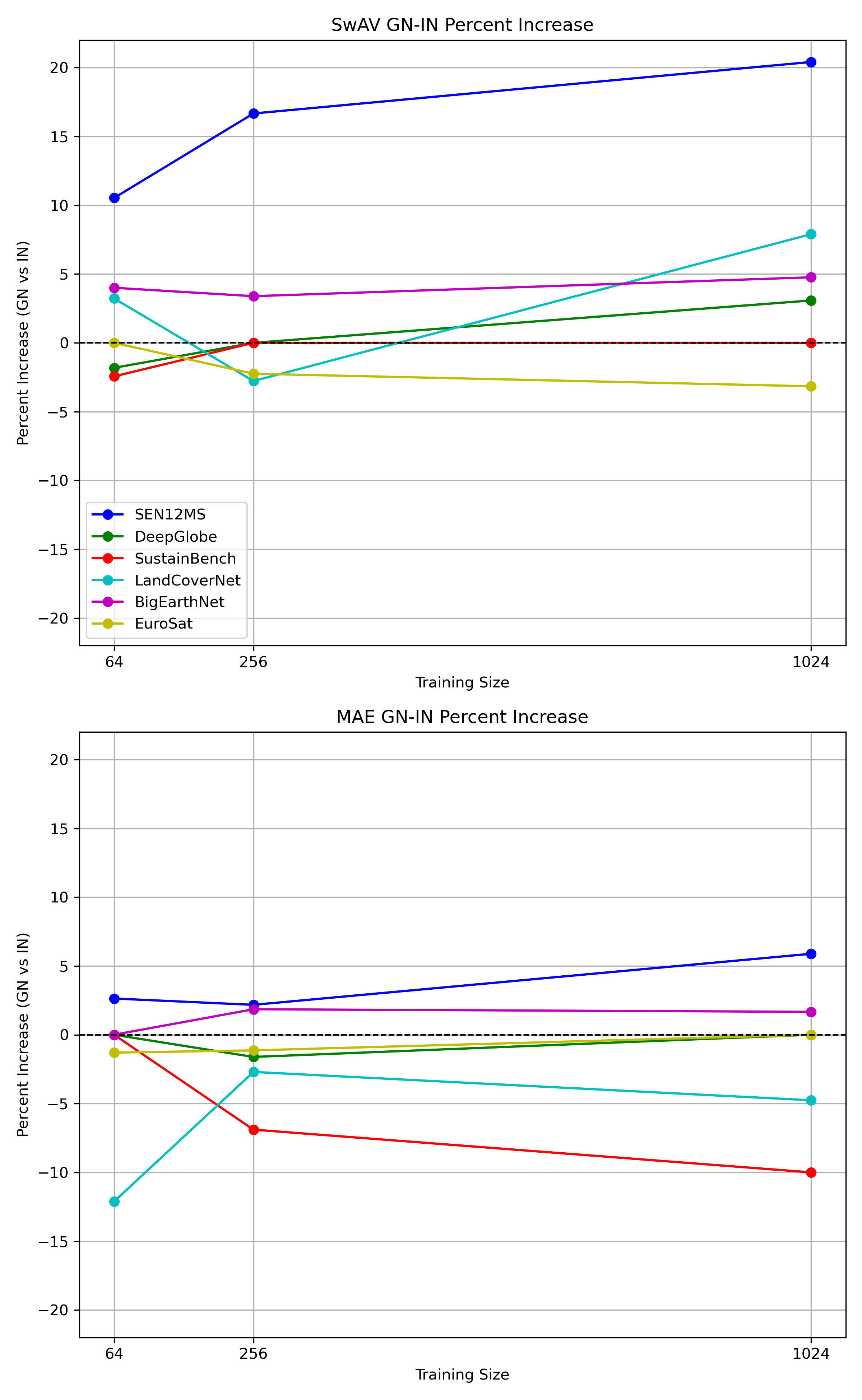}
    \caption{Percent increase when using RSP (GeoNet) vs INP for SwAV and MAE when linear probing on our six benchmark datasets}
    \label{fig:pct_inc}
\end{figure}

\paragraph{Linear Probing}  The LP experiments reveal that self-supervised learning generally outperforms the fully supervised baseline across datasets, consistent with prior literature \cite{wang2022self}. We note that the Sup-IN model achieves the best performance just once out of all 36 scenarios, and generally performs substantially worse than the best SSL model on each task.  We also observe, as expected, that model performances consistently increase as the quantity of labeled data increases.  Comparing the performance of MAE and SwAV, we observe that SwAV tends to perform best on some benchmarks (e.g., SEN12MS, SustainBench, BigEarthNet, and EuroSat), while MAE tends to perform best on others (e.g., LandCoverNet and DeepGlobe).  Our key finding is that there appears to be little advantage, if any, to using RSP (with GeoNet) over INP.  We find that the two pre-training datasets tend to yield similar results when the same SSL approach is employed, and few benchmarks even admit a consistent winner.  If we consider only MAE, we see that INP tends to be superior on just one dataset (LandCoverNet) while RSP tends to be superior also on just one dataset (SEN12MS); in both cases, the performance gap is relatively small.  If we only consider SSL with SwAV, a similar pattern emerges between RSP and INP.  This small performance gap can also be seen in \cref{fig:pct_inc}, where we plot the percent increase in downstream accuracy between RSP and INP. 

\paragraph{Fine-tuning} The FT experiments reveal similar insights to those from the LP experiments. We observe that the SSL methods yield consistent improvements over the supervised baseline as well as consistent performance increase as the training size varies from 64 to 1024 samples. Another shared observation with the LP results is that RSP does not appear to offer a consistent advantage over INP. Comparing the two SSL models, we find that MAE outperforms SwAV on all tasks except SustainBench and BigEarthNet, implying that MAE benefits more from fine-tuning. This further supports the idea that the optimal SSL approach depends on the nature of the downstream task. Surprisingly, we find that fine-tuning performance is generally lower than that of linear probing. We hypothesize that this is due to the relatively limited quantity of training data available for fine-tuning the models, many of which are segmentation decoders that have a large number of free parameters.

\begin{table}[h!]
\centering
\caption{Few-shot performance of GeoNet (GN) and ImageNet (IN)-pretrained SwAV and MAE using 64, 256, and 1024 samples. The top half of the table shows results when linear probing and the bottom half shows the results when these models are finetuned. We report mIoU for segmentation and F-1 score for classification tasks and consider the supervised baseline (Sup.-IN). Results style: \textbf{best}, \underline{second best}.}
\label{tab:few_shot_lin_prob_ft}
\resizebox{\columnwidth}{!}{%
\begin{tabular}{cclcccc|c}
\toprule
\textbf{} & \textbf{Size} & \textbf{Dataset}       & \textbf{MAE-GN} & \textbf{MAE-IN} & \textbf{SwAV-GN} & \textbf{SwAV-IN} & \textbf{Sup.-IN} \\ 
\midrule
\multirow{18}{*}{LP} & \multirow{6}{*}{64}  & SEN12MS        & \underline{0.39} & 0.38 & \textbf{0.42} & 0.38 & 0.33 \\
                     &                      & DeepGlobe      & \textbf{0.58} & \textbf{0.58} & 0.54 & \underline{0.55} & \underline{0.55} \\
                     &                      & SustainBench   & 0.26 & 0.26 & \underline{0.40} & \textbf{0.41} & 0.38 \\
                     &                      & LandCoverNet   & 0.29 & \textbf{0.33} & \underline{0.32} & 0.31 & 0.28 \\
                     &                      & BigEarthNet    & 0.29 & 0.29 & \textbf{0.52} & \underline{0.50} & 0.48 \\
                     &                      & EuroSat        & 0.76 & \underline{0.77} & \textbf{0.81} & 0.81 & 0.78 \\
\cmidrule{3-8} 
                    & \multirow{6}{*}{256}  & SEN12MS        & \underline{0.47} & 0.46 & \textbf{0.49} & 0.42 & 0.40 \\ 
                    &                        & DeepGlobe      & \underline{0.61} & \textbf{0.62} & \underline{0.61} & \underline{0.61} & 0.59 \\ 
                    &                        & SustainBench   & 0.27 & 0.29 & \textbf{0.43} & \textbf{0.43} & \underline{0.42} \\ 
                    &                        & LandCoverNet   & \underline{0.36} & \textbf{0.37} & 0.35 & \underline{0.36} & 0.33 \\ 
                    &                        & BigEarthNet    & 0.55 & 0.54 & \textbf{0.61} & \underline{0.59} & 0.54 \\ 
                    &                        & EuroSat        & 0.87 & \underline{0.88} & 0.87 & \textbf{0.89} & 0.87 \\
\cmidrule{3-8} 
                    & \multirow{6}{*}{1024} & SEN12MS        & \underline{0.54} & 0.51 & \textbf{0.59} & 0.49 & 0.46 \\ 
                    &                        & DeepGlobe      & \textbf{0.69} & \textbf{0.69} & \underline{0.67} & 0.66 & 0.63 \\ 
                    &                        & SustainBench   & 0.27 & \underline{0.30} & \textbf{0.45} & \textbf{0.45} & \textbf{0.45} \\ 
                    &                        & LandCoverNet   & 0.40 & \textbf{0.42} & \underline{0.41} & 0.39 & 0.37 \\ 
                    &                        & BigEarthNet    & 0.61 & 0.60 & \textbf{0.66}& \underline{0.63} & 0.59 \\ 
                    &                        & EuroSat        & \underline{0.93} & \underline{0.93} & 0.92 & \textbf{0.94}& 0.91 \\
\midrule
\multirow{18}{*}{FT} & \multirow{6}{*}{64} & SEN12MS        & \textbf{0.40} & 0.37 & \underline{0.39} & \underline{0.39} & 0.33 \\ 
                    &                     & DeepGlobe      & \textbf{0.59} & \underline{0.58} & 0.56 & 0.56 & 0.51 \\ 
                    &                     & SustainBench   & 0.28 & 0.25 & \textbf{0.42} & \textbf{0.42} & \underline{0.41} \\ 
                    &                     & LandCoverNet   & \underline{0.31} & \textbf{0.32} & 0.30 & 0.31 & 0.26 \\ 
                    &                     & BigEarthNet    & 0.33 & 0.34 & \textbf{0.54} & \underline{0.52} & 0.46 \\ 
                    &                     & EuroSat        & 0.74 & \underline{0.78} & \textbf{0.80} & \underline{0.78} & 0.70 \\
\cmidrule{3-8} 
                    & \multirow{6}{*}{256} & SEN12MS        & \textbf{0.48} & \underline{0.46} & 0.45 & \underline{0.46} & 0.38 \\ 
                    &                     & DeepGlobe      & \textbf{0.64} & \underline{0.63} & 0.60 & 0.61 & 0.55 \\ 
                    &                     & SustainBench   & 0.30 & 0.29 & \textbf{0.44} & \textbf{0.44} & \underline{0.43} \\ 
                    &                     & LandCoverNet   & \textbf{0.36} & \textbf{0.36} & 0.34 & \underline{0.35} & 0.27 \\ 
                    &                     & BigEarthNet    & 0.43 & 0.43 & \textbf{0.59} & \underline{0.58} & 0.53 \\ 
                    &                     & EuroSat        & \textbf{0.91} & \textbf{0.91} & \textbf{0.91} & \textbf{0.91} & \underline{0.86} \\
\cmidrule{3-8} 
                    & \multirow{6}{*}{1024} & SEN12MS        & \textbf{0.55} & \underline{0.53} & 0.48 & 0.48 & 0.40 \\ 
                    &                        & DeepGlobe      & \underline{0.70} & \textbf{0.71} & 0.65 & 0.68 & 0.57 \\ 
                    &                        & SustainBench   & 0.29 & 0.31 & \textbf{0.46} & \textbf{0.46} & \underline{0.45} \\ 
                    &                        & LandCoverNet   & \underline{0.41} & \textbf{0.44} & 0.38 & 0.38 & 0.29 \\ 
                    &                        & BigEarthNet    & 0.53 & 0.49 & \textbf{0.63} & \textbf{0.63} & 0.58 \\ 
                    &                        & EuroSat        & \underline{0.95} &\textbf{0.96} & \underline{0.95} & \underline{0.95} & 0.93 \\

\bottomrule
\end{tabular}%
}
\end{table}

\section{Discussion}
\label{sec:discussion}

In this section, we perform several additional analyses in an attempt to explain why RSP does not yield better performance than INP.

\subsection{Impact of GeoNet Composition}
The design of GeoNet (see Sec. \ref{sec:geonet_description}), as well as other SSL datasets, involves a large number of choices about how to sample available data. It is plausible that some of the design choices used to create GeoNet were suboptimal, leading to the similarity in performance between RSP and INP.  In this experiment, we examine this hypothesis by directly comparing GeoNet against SSL4EO \cite{wang2023ssl4eo}, an independently curated, large Sentinel-2 dataset and evaluating on the same downstream tasks. We performed pre-training on SSL4EO with MAE only due to the computational cost of SwAV pre-training (3 days for MAE vs. 15+ days for SwAV), and we denote the resulting model as MAE-SSL4EO.  In  \cref{tab:ssl_rs} we compare the performance of MAE-GN and MAE-SSL4EO on our downstream benchmark tasks using linear probing, for a training set size of 1024.  The results for smaller training set sizes can be found in the Appendix and reveal a similar pattern. The results indicate that the performance of MAE-SSL4EO and MAE-GN are nearly identical, suggesting that the design choices of the two datasets do not have a major impact on their downstream performance.  Furthermore, this suggests that the findings of our study are not likely due to any unique sub-optimal design choices of GeoNet. 

\subsection{Impact of Pre-Training Procedure}
Recent findings from \cite{isprs-archives-XLIII-B3-2022-1399-2022, zhang2022consecutive} found that obtaining benefits from RSP is most likely when using a so-called "domain adaptive" pre-training, in which a two-stage pre-training process is utilized: first pre-training on ImageNet, and then pre-training on a large RS dataset.  We also investigated whether a domain adaptive pre-training approach might yield better results.  To explore this hypothesis, we pre-trained with MAE for 200 epochs on ImageNet, and then pre-trained on GeoNet for another 200 epochs; we denote the resulting model as MAE-IN-GN.  We restricted the full pre-training process to 400 epochs so that the MAE-IN-GN model did not have an advantage due to greater total pre-training time.  \cref{tab:ssl_rs} presents results comparing MAE-IN-GN against MAE-GN for linear probing with 1024 training samples. The results for smaller training set sizes are presented in the supplement, and reveal a similar pattern.  The results show that MAE-IN-GN outperforms the MAE-GN on five of the six downstream benchmarks, however, the performance advantage in each of these cases is modest (usually 1-2\%). This suggests that there may be some advantage to a two-stage training approach, but it is unlikely to confer a significant performance advantage to RSP compared to INP.

\begin{table}[h!]
\centering
\caption{Linear probing performance using 1024 samples. We compare MAE's performance when pretrained on GeoNet (MAE-GN) and ImageNet (MAE-IN). Lower is better, and best result in each row is bolded.}
\label{tab:ssl_rs}
\resizebox{\columnwidth}{!}{%
\begin{tabular}{@{}lccccc@{}}
\toprule
\textbf{Dataset}       & \textbf{MAE-GN} & \textbf{MAE-SSL4EO} & \textbf{MAE-IN-GN} \\ 
\midrule
SEN12MS         & 0.54 & 0.54 & \textbf{0.55} \\
DeepGlobe       & 0.69 & 0.69 & \textbf{0.70} \\
SustainBench    & 0.27 & 0.27 & \textbf{0.29} \\
LandCoverNet    & 0.40 & 0.41 & \textbf{0.42} \\
BigEarthNet     & 0.61 & 0.61 & \textbf{0.62} \\
EuroSat         & 0.93 & \textbf{0.94} & 0.93 \\ 
\bottomrule
\end{tabular}%
}
\end{table}

\subsection{Does the GeoNet model better encode the benchmark data?}
A major premise of SSL pre-training, is that improvements in the SSL task yields an model that better encodes - in some general sense - the underlying data it was trained upon.  In this section, we investigate whether the MAE-GN model exhibits lower reconstruction error (i.e. its SSL loss) of imagery in the downstream benchmark datasets than the MAE-IN model.  Presumably, the MAE-GN model should consistently yield lower reconstruction error, since the downstream tasks are all the same resolution and modality as the GeoNet dataset (except DeepGlobe, which is 0.3m resolution rather than 10m).  \cref{tab:reconstruction_error} presents the reconstruction errors of MAE-IN and MAE-GN on the GeoNet and ImageNet datasets, as well as our six downstream benchmarks. As expected, MAE-GN achieves the lowest reconstruction error on the GeoNet dataset and MAE-IN does so on the ImageNet dataset, confirming that SSL pre-training was successful. Surprisingly, however, we find that MAE-IN achieves the lowest reconstruction error on three of the six downstream tasks, while MAE-GN achieves the lowest error on the other three.  This suggests that the GeoNet dataset may \textit{not} be representative of the downstream tasks, thereby explaining the similar performance of the MAE-GN and MAE-IN pre-trained models.  \cref{tab:few_shot_lin_prob_ft} plots the relative reconstruction accuracy of MAE-GN and MAE-IN (x-axis) against the downstream benchmark performance (y-axis) for all six benchmark tasks.  Although not a large sample size, we observe that there is a correlation between reconstruction accuracy and downstream task performance, as expected.  This suggests that a key to improving RSP may be uncovering why MAE-GN does not provide improved reconstruction of the downstream tasks.  

\begin{table}[h!]
\centering
\caption{Reconstruction error of MAE pretrained on ImageNet, GeoNet, both, and SSL4EO on min(N, 100k) samples from the pretraining datasets and benchmark tasks, where N is the dataset size. Lower is better. Results style: \textbf{best}}
\label{tab:reconstruction_error}
\begin{tabular}{@{}lccccc@{}}
\toprule
\textbf{Dataset}       & \textbf{MAE-GN} & \textbf{MAE-IN}   \\ 
\midrule
GeoNet                  & \textbf{0.083} & 0.097  \\
ImageNet                & 0.327          & \textbf{0.251}  \\
SEN12MS                 & \textbf{0.079} & 0.081            \\
DeepGlobe               & 0.218          & \textbf{0.211}   \\
SustainBench            & 0.164          & \textbf{0.159}   \\
LandCoverNet            & 0.098          & \textbf{0.076}   \\
BigEarthNet             & \textbf{0.095} & 0.106         \\
EuroSAT                 & \textbf{0.095} & 0.112         \\
\bottomrule
\end{tabular}%
\end{table}

\begin{figure}
    \centering
    \includegraphics[width=\linewidth]{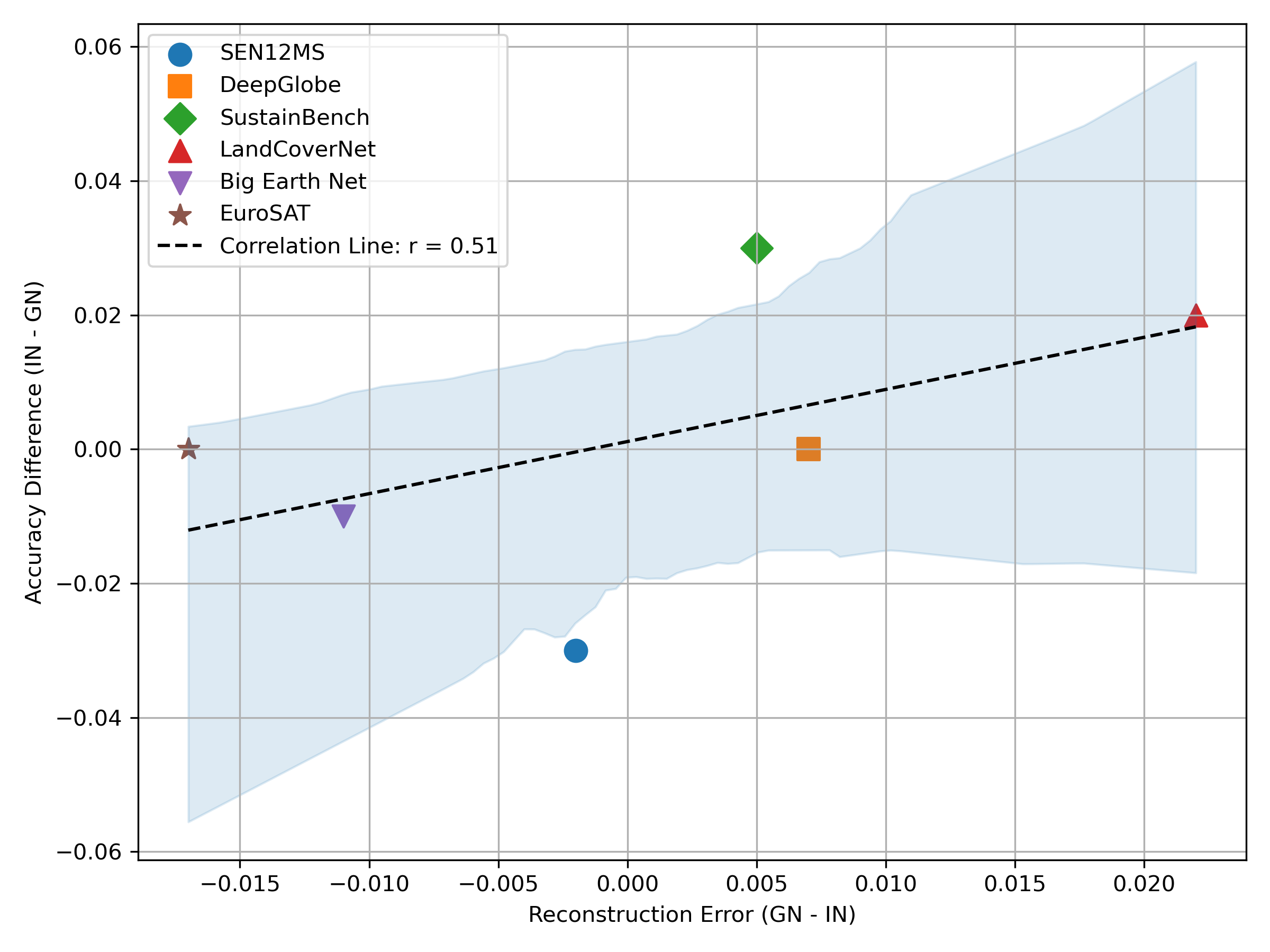}
    \caption{Difference in reconstruction error and in downstream performance are weakly positively correlated. We plot the difference between reconstruction error between MAE-GN and MAE-IN and the accuracy difference between MAE-IN and MAE-GN for linear probed models using 1024 training samples. The dotted regression line describe the correlation, and the 95\% confidence intervals is shaded. The Pearson’s correlation coefficient is shown in the legend.}
    \label{fig:corr_mae}
\end{figure}

\section{Conclusions}
\label{sec:conclusions}
 In this work, we investigated whether pre-training on remote sensing data with state-of-the-art self-supervised learning (SSL) techniques improves downstream performance on remote sensing recognition tasks. To address this question, we constructed a large and diverse dataset of optical satellite imagery, termed GeoNet.  We pre-trained models on GeoNet and ImageNet, respectively, using two different SSL pre-training strategies: SwAV and MAE. The resulting four models were fine-tuned on six downstream tasks with ground truth labels. \textbf{Surprisingly, the results showed no consistent advantage to pre-training with GeoNet as compared to ImageNet, regardless of whether SwAV or MAE was used.} Subsequent analysis revealed that neither variations in the construction of the GeoNet dataset, nor using a more complex two-stage "consecutive pre-training" strategy altered these findings.  
 
 \paragraph{Limitations} We highlight two key limitations of our study. First, although Sentinel-2 imagery includes thirteen total bands, we restricted GeoNet to use only the RGB bands. Our motive for this choice was twofold: (i) by using RGB imagery, existing state-of-the-art SSL methods used on ImageNet are more transferrable, without requiring modifications; and (ii) we wished to focus our study on whether INP is sufficient for downstream tasks involving RGB remotely-sensed data, which is a common use case for SSL in remote sensing.  We note that \cite{hakizimana2024enhanced} has also provided a preliminary investigation of SSL pretraining on ten Sentinel-2 bands. Nonetheless, the additional spectral bands remain relevant for many downstream tasks using Sentinel-2, making their inclusion an important opportunity for future work.

 A second notable limitation pertains to our experimental design. As detailed in \cref{sec:experiments}, we control many aspects of our experimental design to ensure a fair comparison between INP and RSP.However, the INP imagery fed into pre-training is sampled from a larger overall pool of pixels than RSP - about four times larger - providing a potential advantage to INP. We presented arguments in \cref{sec:experiments} that the effective increase in pixels is much smaller than four, however it may still impart an advantage to INP, and therefore represents an important limitation.

\section*{Acknowledgments}
This research was supported by the Duke University Bass Connections program and the Duke University Climate+ program.

\bibliography{egbib}\bibliographystyle{ieee_fullname}

\newpage

\appendix

\section{GeoNet Sampling}
\label{supp:sampling}

The coordinates for extracting Sentinel-2 images are collected in two rounds. During the first sampling round, 1.5 million coordinates are generated using the four sampling methods outlined the main paper. We identify the corresponding grid cells containing these coordinates and limit the number of images per grid cell to a maximum of four. Samples exceeding this threshold are replaced with 'budget' samples generated using method 1 (urban \& rural), maintaining the urban-to-rural ratio of 2:1. For grid cells sampled more than once, we extract images from different search periods to introduce seasonal diversity and reduce redundancy, using 16 seasons (3 month periods) between January 2019 and January 2023.  

\begin{table}[h]
\centering
\caption{Built Features Datasets Used}
\label{tab:built_features_datasets}
\resizebox{\columnwidth}{!}{%
\begin{tabular}{p{2.2cm}p{5cm}p{1cm}}
\toprule
\textbf{Category} & \textbf{Dataset} & \textbf{Total} \\
\midrule
Airport         & The Global Airport Database (GADB)       & 4K   \\
Mining Sites    & Global-scale mining polygons            & 20K  \\
Power Plants    & Global Power Plant Database             & 35K  \\
Ports           & The World Port Index (Pub 150)          & 3K   \\
Oil Rigs        & Global Oil and Gas Extraction Tracker   & 9K   \\
Wind Turbines   & DeepOWT: a global offshore wind turbine data set & 5K   \\
Dams            & The World Bank Global Dams Database     & 6K   \\
\bottomrule
\end{tabular}
}
\end{table}

\begin{table}[h]
\centering
\caption{Auxiliary datasets and weight proxy variables for sampling natural disasters}
\label{tab:auxiliary_datasets}
\resizebox{\columnwidth}{!}{%
\begin{tabular}{p{2.2cm}p{4cm}p{4cm}p{1cm}}
\toprule
\textbf{Disaster} & \textbf{Dataset} & \textbf{Proxy Variable} & \textbf{Total} \\
\midrule
Wildfire & MODIS Active Fire Data & Brightness temperature measured (in Kelvin) of observed fire hotspot & 0.4M \\
Flood    & Global Flood Mortality Risks and Distribution, v1 (2000) & Estimated flood-based mortality risks & 2.5M \\
Drought  & Global Drought Mortality Risks and Distribution & Mortality loss estimates per drought event & 0.6M \\
Cyclone  & Global Cyclone Hazard Frequency and Distribution & Relative frequency of cyclone hazard & 0.5M \\
\bottomrule
\end{tabular}
}
\end{table}

\begin{figure}[h]
    \centering
    \includegraphics[width=\linewidth]{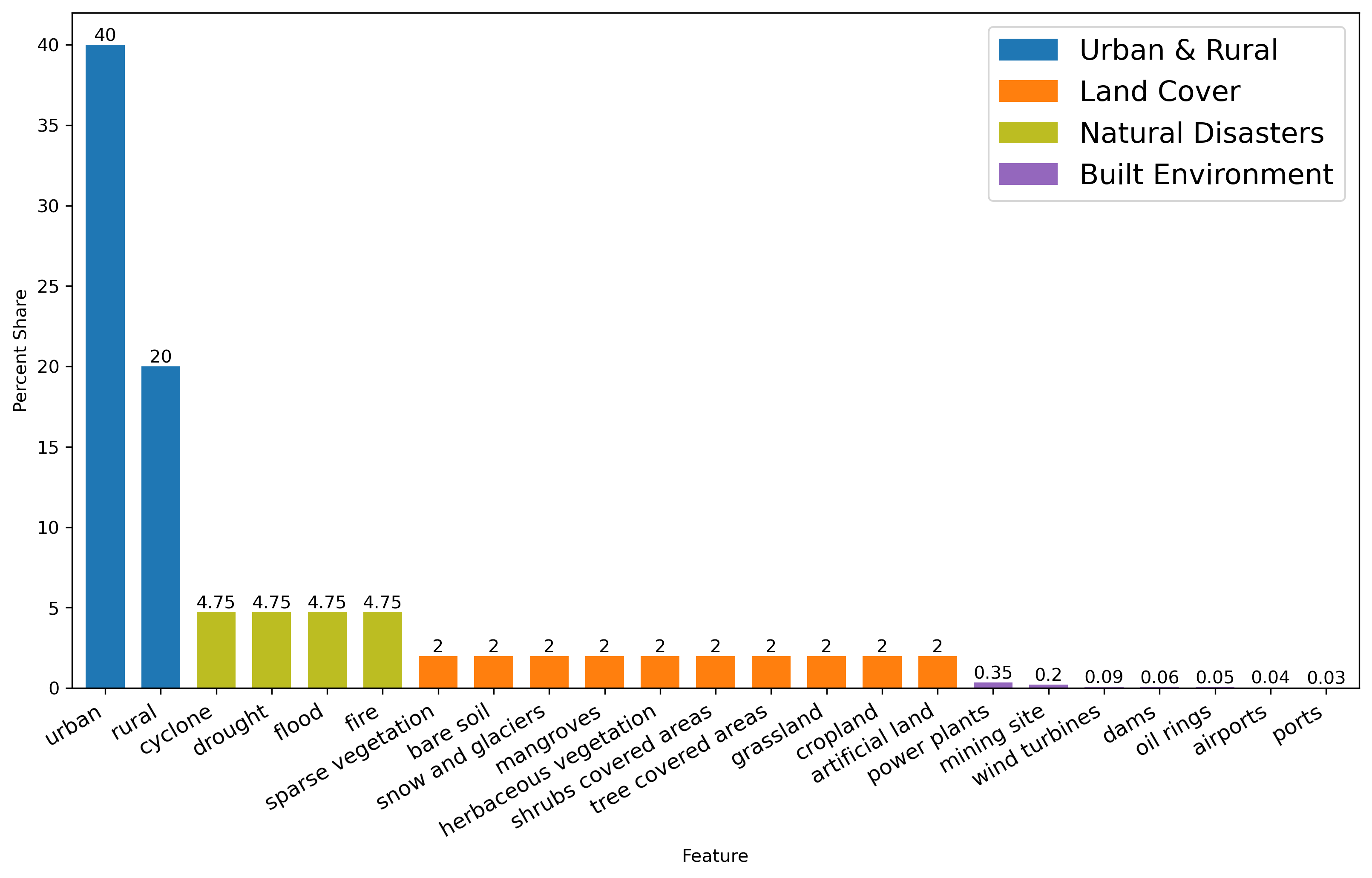}
    \caption{Percentage breakdown of our four sampling methods and features used to collect GeoNet}
    \label{fig:label_dist}
\end{figure}
We describe the auxiliary datasets used to sample locations for built features and natural disasters. \cref{tab:built_features_datasets} shows the built features datasets used and their total observations. \cref{tab:auxiliary_datasets} presents the datasets used to samples areas of natural disasters and the proxy variable used for each. 

\section{GeoNet Processing}

We use Google Earth Engine (GEE) to download optical Sentinel-2 images for the 2.24km x 2.24km patches around the grid cells collected (code available in \hyperlink{https://github.com/zcalhoun/data-plus-22/blob/main/imageExporter.py}{this repository}). For Sentinel-2, we use GEE's "ee.Image.visualize" method to produce RGB images using the B4, B3, and B2 bands (corresponding to Red, Green, and Blue, respectively), and save them in GeoTIFF format. Images with a cloud percentage higher than 10\% are filtered out following the threshold established by \cite{manas2021seasonal}. We remove empty or corrupt images, identified by their small file sizes. Our final dataset size is 1,228,895, which is close to the size of ImageNet-1k (1,281,167), including 1,128,070  unique locations.

\section{Experiments}

\subsection{Pre-training experimental design details}
Following \cite{caron2020unsupervised}, for SwAV we train a ResNet-50 for 400 epochs using a batch size of 256 on 8 NVIDIA RTX 3090 GPUs\footnote{To fit on our GPUs, we limit the SwAV experiments to a batch size of 256, which was shown in \cite{caron2020unsupervised} to be competitive with its 4096 counterpart.}, and we denote the resulting pre-trained models as SwAV-IN and SwAV-GN, respectively.  Following \cite{he2022masked}, for MAE we train a ViT-B for 400 epochs and a batch size of 4096 on 8 NVIDIA RTX Titan GPUs, and denote the resulting pre-trained models as MAE-IN and MAE-GN, respectively. We also include a ResNet50 model with supervised pre-training as a baseline. We follow default pre-processing approaches for both models, as per the original publications \cite{caron2020unsupervised, he2022masked} (MAE: RandomResizedCrop, RandomHorizontalFlip, Normalization; SwAV: same as MAE as well as color distortions and RandomGaussianBlur) for both GeoNet and ImageNet. Specifically, both SwAV and MAE use RandomResizedCrop with size 224 and a scale range of (0.14-1) for SwAV and (0.2-1) for MAE. Consequently, given ImageNet images are larger on average ((473, 405) vs. (224, 224)), they contain about four times more unique pixels than GeoNet.

\subsection{Benchmark Datasets}
\label{sub:benchmark_datasets}

\textbf{SEN12MS (seg.)} \cite{schmitt2019sen12ms} provides 180,662 triplets of Sentinel-1 and Sentinel-2 images at 256×256 pixels and 10 m/px resolution, covering diverse global regions across all four meteorological seasons. The dataset uses “weak labels” derived from MODIS land cover maps. Weak labels provide probabilistic estimates for each label in the dataset. We only use the LCCS surface hydrology schema \cite{di2005land} as it has the highest reported accuracy among the four schemas (87\%). The labels include 10 classes: dense forest, open forest, shrubland, grasslands, herbaceous wetlands, herbaceous croplands, tundra, permanent snow and ice, barren, and water bodies.  We center-crop the images to 224x224 pixels and use stratified samples by season to create our train/val/test splits.

\textbf{DeepGlobe Land Cover (seg.)} \cite{demir2018deepglobe} consists of 80,301 satellite images captured by DigitalGlobe’s satellite at 0.5m/px resolution. The dataset includes seven land cover annotation classes (urban land, agriculture land, rangeland, forest land, water, barren land, and unknown) collected over Thailand, Indonesia, and India. It is split into training, testing, and validation sets in a 70/15/15 ratio. The original images, sized 2448×2448, (along with their labels), are divided into 224×224 tiles, producing 100 tiles per image. Pixels at the margins are excluded to maintain a consistent tile size. We ignore the unknown class for loss and accuracy computations. 

\textbf{Farm Parcel Delineation (seg.)} \cite{aung2020farm} focuses on the delineation of farm parcel boundaries in France using Sentinel-2 RGB images and binary labels at 224×224 pixels with a 10 m/px resolution. No additional preprocessing is needed: we use the train/val/test split proposed in \cite{yeh2021sustainbench} (80/10/10), which ensures independence among the sets. The final dataset consists of 1572 training, 198 validation, and 196 test samples.

\textbf{BigEarthNet (multi-label cls.)} \cite{sumbul2019bigearthnet} is a large benchmark dataset containing 590,326 pairs of Sentinel-1 and Sentinel-2 image patches across 10 European countries. The 10 m/px images come in 120×120, which we resize to 224x224 to match the rest of our datasets, and following recommendations of \cite{corley2024revisiting}. The images are annotated with 43 classes based on the CORINE Land Cover database. The dataset supports multi-label classification tasks, and is structured using a list of subdirectories each containing a satellite image and a .json metadata file. We use the .json files to create a one-hot encoding multilabel .csv file, which we use to perform stratified sampling by land cover class to create our train/val/test splits.

\textbf{EuroSat (multi-class cls.)} \cite{helber2019eurosat} consists of 27,000 Sentinel-2 images with 10 different land use and land cover classes (including residential, industrial, and various types of vegetation). Each image is 64x64 pixels at a resolution of 10m/px. To match the rest of the benchmarks, we upsample each image to 224x224 pixels using pixel area relation. We split the dataset into training, validation, and testing datasets using a 65/15/20 percentages to match the 80/20 training-test split suggested in the original paper. We conduct stratified sampling to keep the representation of each class the same during finetuning. 

\textbf{LandCoverNet (seg.)} \cite{alemohammad2020landcovernet} includes 1,980 Sentinel-2 images and their land cover labels across Africa. Using a consensus method, three human annotators assign soft labels for each land cover type: no data, water, snow/ice, bare ground (artificial or natural), and vegetation (woody or non-woody, cultivated or semi-natural). All images are 256x256 pixels with a 10 m/px resolution. We center crop the images to 224x224 and ignore the no data class for loss and accuracy computations. 

\section{Additional Results}
\label{app:results}
In \cref{tab:few_shot_lin_prob_ft_others}, we summarize the results obtained by linear probing and fine-tuning models pre-trained on SSL4EO and using model-adaptive pre-training (i.e. first on ImageNet, and subsequently on GeoNet) across training sizes and benchmark tasks.

\begin{table}[h]
\centering
\caption{Few-shot performance of SSL4EO and Domain-Adaptive (IN-GN) pretrained MAE using 64, 256, and 1024 samples. The top half of the table shows results when linear probing and the bottom half shows the results when these models are finetuned. We report mIoU for segmentation and F-1 score for classification tasks and consider the supervised baseline (Sup.-IN).}
\label{tab:few_shot_lin_prob_ft_others}
\resizebox{\columnwidth}{!}{%
\begin{tabular}{cclcc}
\toprule
\textbf{} & \textbf{Size} & \textbf{Dataset} & \textbf{MAE-SSL4EO} & \textbf{MAE-IN-GN} \\ 
\midrule
\multirow{18}{*}{LP} & \multirow{6}{*}{64}  & SEN12MS        & 0.39 & 0.39  \\
                     &                      & DeepGlobe      & 0.58 & 0.59 \\
                     &                      & SustainBench   & 0.25 & 0.26   \\
                     &                      & LandCoverNet   & 0.31 & 0.32 \\
                     &                      & BigEarthNet    & 0.29 & 0.30 \\
                     &                      & EuroSat        & 0.78 & 0.76 \\
\cmidrule{3-5} 
                    & \multirow{6}{*}{256}  & SEN12MS        & 0.48 & 0.49  \\
                     &                      & DeepGlobe      & 0.61 & 0.63 \\
                     &                      & SustainBench   & 0.25 & 0.30   \\
                     &                      & LandCoverNet   & 0.35 & 0.37 \\
                     &                      & BigEarthNet    & 0.55 & 0.56 \\
                     &                      & EuroSat        & 0.88 & 0.87 \\
\cmidrule{3-5} 
                    & \multirow{6}{*}{1024}  & SEN12MS        & 0.54 & 0.55  \\
                     &                      & DeepGlobe      & 0.69 & 0.70 \\
                     &                      & SustainBench   & 0.27 & 0.29   \\
                     &                      & LandCoverNet   & 0.41 & 0.42 \\
                     &                      & BigEarthNet    & 0.61 & 0.62 \\
                     &                      & EuroSat        & 0.94 & 0.93 \\
\midrule
\multirow{18}{*}{FT} & \multirow{6}{*}{64} & SEN12MS        & 0.40 & 0.40 \\ 
                     &                     & DeepGlobe       & 0.59 & 0.59 \\ 
                     &                     & SustainBench    & 0.25 & 0.28 \\ 
                     &                     & LandCoverNet    & 0.31 & 0.32 \\ 
                     &                     & BigEarthNet     & 0.35 & 0.32 \\ 
                     &                     & EuroSat         & 0.75 & 0.75 \\
\cmidrule{3-5} 
                     & \multirow{6}{*}{256} & SEN12MS        & 0.48 & 0.48 \\ 
                     &                     & DeepGlobe       & 0.62 & 0.64 \\ 
                     &                     & SustainBench    & 0.27 & 0.30 \\ 
                     &                     & LandCoverNet    & 0.35 & 0.38 \\ 
                     &                     & BigEarthNet     & 0.44 & 0.42 \\ 
                     &                     & EuroSat         & 0.90 & 0.92 \\
\cmidrule{3-5} 
                    & \multirow{6}{*}{1024} & SEN12MS        & 0.54 & 0.56 \\ 
                    &                     & DeepGlobe       & 0.70 & 0.72 \\ 
                    &                     & SustainBench    & 0.28 & 0.31 \\ 
                    &                     & LandCoverNet    & 0.41 & 0.43 \\ 
                    &                     & BigEarthNet     & 0.52 & 0.50 \\ 
                    &                     & EuroSat         & 0.95 & 0.96 \\

\bottomrule
\end{tabular}%
}
\end{table}

\end{document}